# Bypass Enhancement RGB Stream Model for Pedestrian Action Recognition of Autonomous Vehicles

Dong Cao[1,2] and Lisha Xu[1,2]

[1] Institute of Cognitive Intelligence, DeepBlue Academy of Sciences
[2] DeepBlue Technology (Shanghai) Co., Ltd. No.369, Weining Road, Shanghai, China
doocao@gmail.com, xuls@deepblueai.com

**Abstract.** Pedestrian action recognition and intention prediction is one of the core issues in the field of autonomous driving. In this research field, action recognition is one of the key technologies. A large number of scholars have done a lot of work to improve the accuracy of the algorithm for the task. However, there are relatively few studies and improvements in the computational complexity of algorithms and system real-time. In the autonomous driving application scenario, the real-time performance and ultra-low latency of the algorithm are extremely important evaluation indicators, which are directly related to the availability and safety of the autonomous driving system. To this end, we construct a bypass enhanced RGB flow model, which combines the previous two-branch algorithm to extract RGB feature information and optical flow feature information respectively. In the training phase, the two branches are merged by distillation method, and the bypass enhancement is combined in the inference phase to ensure accuracy. The real-time behavior of the behavior recognition algorithm is significantly improved on the premise that the accuracy does not decrease. Experiments confirm the superiority and effectiveness of our algorithm.

## 1 Introduction

In the field of autonomous vehicles, pedestrian action recognition and intention prediction are one of the core issues to be solved urgently, which directly affects the process of autonomous vehicles to a higher level. In the field of pedestrian action recognition and intention prediction, we can start from a variety of perspectives. Behavior recognition is an important part. This paper studies how to improve the real-time behavior recognition algorithm under the premise of ensuring accuracy.

Visual understanding is one of the core issues of artificial intelligence, and it has been rapidly developed with the favorable promotion of deep learning technology. As an important direction of visual understanding, action recognition is the basic work for further application. Current methods to deal with the problem almost following three ways: a) Two-stream frameworks that consider spatial and temporal information by taking RGB and optical flows as input[1]; b) 3D Network that use 3D convolutional kernel to extract the spatial and temporal features simultaneously[2]; c) CNN+RNN that process the visual input with a CNN whose outputs are fed into a stack of RNN (LSTM is common)[3].

These approaches above all prove that the motion information in video plays an essential role in action recognition. As a typical representation of motion information, optical flow is calculated as a drift in a short time, meaning the moment velocity. Moreover, 3D spatiotemporal CNN also found that RGB + optical flow boost their accuracy, and achieve the state-of-the-art result[4] in UCF101[5] and HDMI51 datasets[6].

There have been some attempts to describe the optical flow. Dense trajectories track the feature points of frames based on displacement information from optical flow fields, then train the classifiers with the encoded motion features extracted from trajectories[7]. IDT improves the optical flow image by eliminating the camera motion and optimizing the normalization, showing a superior stability but unsatisfactory speed[8]. TV-L1 method is appealing for its good balance between accuracy and efficiency, which iteratively calculates the displacements[9].

Since optical algorithms mentioned above are offline, Flownet was proposed, which is end-to-end trainable, including a generic architecture and another one containing a layer that correlates feature vectors at different locations of image, enabling the online predication of optical flow[10]. Regarding to the quality of flow, Flownet2.0 fuses a stacked network with small displacement network in an optical manner, resulting in a great balance between accuracy and speed on real-world datasets[11]. However, although they were superior in terms of the accuracy, they suffered from extremely expensive computation in terms of both time and storage.

In this study, we propose a novel architecture Bypass Enhancement RGB Stream Model, which leverages the prior information of complex model to obtain the model parameters of motion information during training, and processes RGB information through extended branches in the backbone. This model reduces the high computational consumption caused by optical flow, dynamically adjusts the ratio of RGB information to motion information, and avoids over-reliance on optical flow information when using the same ratio to process different dynamic video, such as the traditional two-stream model to infer static videos.

## 2 Related Work

**Hallucination** Since the computation of optical flow is time consuming and storage demanding, some attempts to learn other way to replace the flow to represent motion information. (Wu, C.Y.et al, 2017) [12]proposed that compressed video algorithms can decrease the redundant information, so that can be used accumulated motion vector and residuals to describe motion. Compared to traditional flow methods, motion vectors bring more than 20 times acceleration although a significant drop in accuracy. Some methods represent motion information only by RGB frame[13, 14]. (Gao, Ruohan.et al, 2018) [15]considered that a static image can produce fake motion, thus predict optical flow fields through pre-trained Im2Flow network. [16]used to hallucinate optical flow images from videos. Monet models the temporal relationship of appearance features and concurrently calculates the contextual relationship of optical flow features using LSTM[17]. All these approaches describe motion information implicitly but achieve good performance in action recognition.

**Generalized Distillation** As a method of model compression[18], knowledge distillation can leverage a more complex model that has been trained to guide a lighter model training, so as to keep the original accuracy of large model while reducing the model size and computing resources[19, 20]. (Geoffrey Hinton et al, 2015)[20]minimized the linear combination of loss functions that are cross entropy with the soft targets and cross entropy with the correct labels. Human society have proved experimentally that interactions between teacher and student can significantly accelerate the learning process. According to this theory, [21]considered a model that supplies privileged information by paradigms when training. [22] applied the privileged information to complex model during distillation to finish knowledge transfer. Recently, for its superior performance in supervised, semi-supervised and multitask learning tasks, more works derived from distillation and privileged knowledge look forward to improve their tasks[23, 24, 25, 26, 27].

## 3    Our approach

Two methods were proposed in literature [28], namely MERS and MARS. MARS realizes the information fusion between the optical flow branch and the RGB branch by constructing a two-part loss function. In the phase of model training, it is still necessary to calculate the optical flow and extract the optical flow feature, and then realize feature information transmission from the optical flow feature to the RGB feature branch through the distillation method. In the inference phase, we only use the RGB model to complete the recognition task. It is not necessary to use the optical flow branch, avoiding the calculation of the optical flow. Inference phase can significantly reduce the computational complexity and improve the real-time performance. However, it is at the expense of proper sacrifice accuracy.

In order to solve the problem of this method in [28], we construct a Bypass Enhancement RGB Stream (BERS) Model. It is hoped that the performance of accuracy can be ensured under the condition that the computational complexity is reduced, and the model can infer in real time.

The BERS model structure is shown in Figure 1. The performer of the model is divided into two different operating states: training mode and inference mode. The overall structure was inspired by[28] . The overall structure of the figure includes upper and bottom parts. In the training mode, the RGB frames of the video are input into model. RGB information is first sent to the bottom model, and the optical flow frames are processed by the optical flow algorithm module, and then the optical flow based action recognition model training is performed. Training is completed to obtain the learned feature weights. Then the global model training is performed, the input is still the RGB frames of the video, and the bottom feature weights are used for the distillation algorithm to assist in learning of the upper model. In the inference mode, the input is the RGB frames of the video, and only the upper model is activated to implement the inference, and the bottom model does not work. The detailed description is as follows:

**Training phase**

In the training mode, the bottom model and the upper model shown in Figure 1 are all involved in the operation, but the participating operations are sequential and not involved in the operation at the same time.

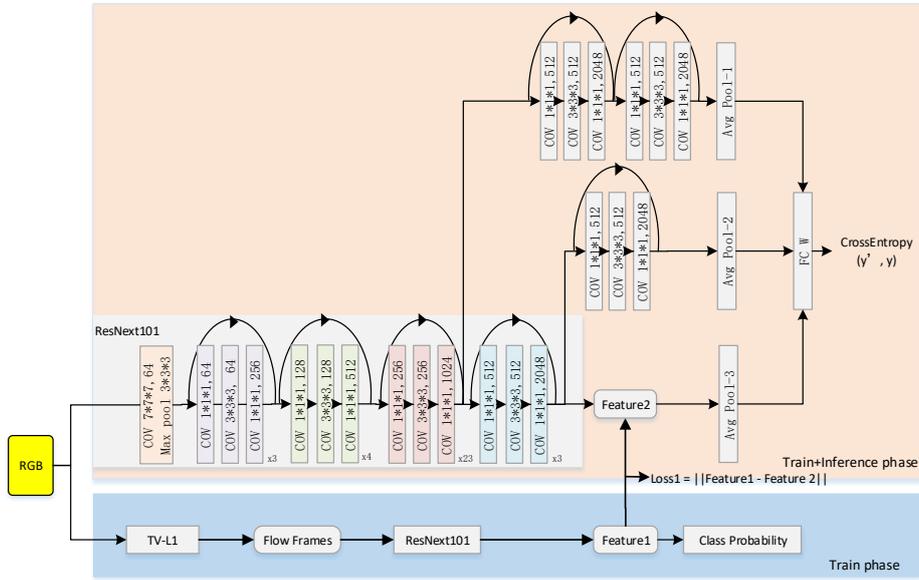

**Fig1.** Bypass Enhancement RGB Stream Model. The bottom is the optical flow branch in training phase, the upper is the enhancement RGB branch.

The first is the training of the bottom model. The role of the bottom model is to extract and learn the optical flow information of the video. The input is RGB frames, which pass through the optical flow algorithm module. Here we use the TV-L1 algorithm to obtain optical flow frames. The optical flow frames are further sent to the deep convolutional neural network for optical flow feature information extraction. Here, the network uses 3D ResNeXt101[31, 32], and the output Feature1 of optical flow feature is connected to Class Probability. Then the loss function is constructed according to the cross entropy of Class Probability and the label y, optimized iteratively to completes the model parameter learning. After the training of bottom model is completed, the valuable material we need to use is the optical flow feature Feature1.

Next, it is the training of the upper model. In this training phase, we need to assist with the bottom model that has been pre-trained. Specifically, the input original RGB frames are simultaneously input to the bottom model and the upper model, respectively.

When the RGB video frames enter into the already trained model in bottom, the parameters of the bottom model are fixed, and then the optical flow feature Feature1 corresponding to the input RGB frames can be obtained. At the same time, the original RGB frames are input into the upper model, first enter into the deep convolutional neural network to extract the apparent spatial feature information, and then are divided into three branches to construct the first part of the loss function.

The first branch is output to a small residual network-1, and then to average pool layer 1, where the input of the first branch is taken from the penultimate ResNext block of deep network 3D ResNeXt101; the second branch is output to another small residual network-2, and then to average pool layer 2, where the input point of the second branch is taken between the last convolution layer of the deep network 3D ResNeXt101 and Feature2; the third branch is output to Feature2, then to average pool layer 3; then connect the above three outputs from Avg pool-1, Avg pool-2, and Avg pool-3 through the fully connected layer, and construct the first part of the loss function with the label y.

$$La = CrossEntropy[W(Avg\ poll-1、Avg\ poll-2、Avg\ poll-3), y] \quad (1)$$

Where W stands for the weights of the fully connected layer.

The second part of the loss function adopts a similar method in[28], and uses the distillation method to realize the transmission of optical flow frame information to the upper model, called Loss 1. The overall form of the loss function is:

$$L = La + Loss1 = La + \lambda ||Feature1 - Feature2|| \quad (2)$$

Where $\lambda$ is a hyper-parameter that used to adjust the effect of distillation.

The training of the upper model is completed under the above loss function condition.

**Inference phase**

In the inference mode, only the upper model is used, and the bottom model does not work. After the video needed to inference are given, the model outputs the action category.

## 4    Experiment

**Dataset** In this section, we will investigate the performance of the Bypass Enhancement RGB Stream framework on datasets Kinetics400[29] and UCF101[30]. Kinetics400 consists of 400 classes, including 246k training videos, 20k validation videos, and each video are from YouTube, about 10 seconds. Kinetics is a large-scale dataset whose role in video understanding is somewhat similar to ImageNet's role in image recognition. In this paper, we use the validation dataset to test our trained model. Some work migrates to other video datasets using the Kinetics pre-training model. UCF101 is a dataset containing 101 classes that belong to five categories: makeup brushing, crawling, haircutting, playing instruments, and sports. 25 people perform each type of action. Following setting of [28], we use the split1 during training and average 3 splits when testing.

**Implementation Details** The novel architecture is composed with source optical flow branch and target derived RGB branch. According to the spirit of distillation and privileged knowledge, when training, we use TV-L1 method [9] to extract optical frames, and save them in jpeg format as the input for complex model. Due to the performance of ResNext101[31, 32] architecture, we adopt it to extract the features after inputting the RGB and optical flows. For the derived branch of RGB, we choose improved ResNext. Following the setting of [11, 28], we train the model with the SGD optimization method, and use 64 frames clips both in training and testing. As for the weights of loss, grid-search is applied to find the important hyper-parameters. We train the model on Kinetics from scratch while finetune from the pre-trained Kinetics400 model on UCF101.

**Results** This part illustrates the superiority of our model from three perspectives. First we compare our model with some single stream models and two-stream models. According to Table 1, we can see that motion information is more accurate than RGB information basically. And two-stream models perform better than single ones. Bypass Enhancement RGB Stream Model is 3.5% higher than MARS while 0.9% lower than MARS+RGB on Kinetics dataset, and 0.9% higher than MARS while 0.1% lower than MARS+RGB on UCF101 dataset. Second in the case of videos that recognize static actions, our model standout(see Table 2). Third we compare our model with some state-of-the-art models(see Table 3). Apparently our model maintains a good accuracy while reducing a large amount of computing resources.

**Table 1.** Single-stream model and two-stream models(Dataset: validation of Kinetics and split 1 of UCF101).

| Stream | Kinetics | UCF101-1 |
| --- | --- | --- |
| RGB | 68.2% | 91.7% |
| Flow | 54.0% | 92.5% |
| MERS | 54.3% | 93.4% |
| MARS | 65.2% | 94.6% |
| RGB+Flow | 69.1% | 95.6% |
| MERS+RGB | 68.3% | 95.6% |
| MARS+RGB | 69.6% | 95.6% |
| OUR | 68.7% | 95.5% |

**Table 2.** Recognition on video cases with static actions

| Action | MARS | OUR |
| --- | --- | --- |
| Making sushi | 24% | 35.2% |
| Eating cake | 4% | 14.1% |
| Reading newspaper | 6% | 17.7% |

**Table 3.** Compare with state-of-the-art models(Dataset: validation of Kinetics and average 3 splits of UCF101)

| Model | Kinetics | UCF101 |
|---|---|---|
| Two-stream | 69.1% | 88.0% |
| ResNext101 | 65.1% | 94.5% |
| I3D | 71.1% | 98.0% |
| MARS+RGB+FLOW | 74.9% | 98.1% |
| OUR | 68.7% | 97.2% |

## 5    Conclusion

In this paper, we propose a novel model, named Bypass Enhancement RGB Stream Model, to distill the motion information from a complex model during training, avoid expensive computation by only taking RGB images as input when testing. This model combine the appearance features and motion feature through a linear combination of losses, resulting a good balance of accuracy and time in dataset Kinetics and UCF101.